\documentclass[lettersize,journal]{IEEEtran}
\usepackage{amsmath,amsfonts}
\usepackage{algorithmic}
\usepackage{algorithm}
\usepackage{array}
\usepackage[caption=false,font=normalsize,labelfont=sf,textfont=sf]{subfig}
\usepackage{textcomp}
\usepackage{stfloats}
\usepackage{url}
\usepackage{verbatim}
\usepackage{graphicx}
\usepackage{cite}
\usepackage{booktabs}
\usepackage{xcolor}

\newcommand{\ours}{{Con2EM}}
\newcommand{\citeay}[1]{(\citeyear{#1})}
\renewcommand{\citeay}[2]{(#2, \cite{#1})}
\newtheorem{theorem}{Theorem}

\begin{document}

\title{Guidance Not Obstruction: A Conjugate Consistent Enhanced Strategy for Domain Generalization}

\author{Meng Cao, Songcan Chen
\thanks{Meng Cao and Songcan Chen are with the MIIT Key Laboratory of Pattern Analysis and Machine Intelligence, Nanjing University of Aeronautics and Astronautics, Nanjing 210016, China, and also with the College of Computer Science and Technology, Nanjing University of Aeronautics and Astronautics, Nanjing 210016, China (e-mail: meng.cao@nuaa.edu.cn, s.chen@nuaa.edu.cn).}
\thanks{\textit{Corresponding author: Songcan Chen}}
\thanks{Manuscript received April 19, 2021; revised August 16, 2021.}}

\markboth{Journal of \LaTeX\ Class Files,~Vol.~14, No.~8, August~2021}%
{Shell \MakeLowercase{\textit{et al.}}: A Sample Article Using IEEEtran.cls for IEEE Journals}

\IEEEpubid{}

\maketitle

\begin{abstract}
Domain generalization addresses \textit{domain shift} in real-world applications.
Most approaches adopt \textit{a domain angle}, seeking invariant representation across domains by aligning their marginal distributions, irrespective of individual classes, naturally leading to insufficient exploration of discriminative information.
Switching to \textit{a class angle}, we find that multiple domain-related peaks or clusters within the same individual classes must emerge due to distribution shift.
In other words, marginal alignment does not guarantee conditional alignment, leading to suboptimal generalization. 
Therefore, we argue that acquiring discriminative generalization between classes within domains is crucial.
In contrast to seeking distribution alignment, we endeavor to safeguard domain-related between-class discrimination.
To this end, we devise a novel Conjugate Consistent Enhanced Module, namely \ours, based on a distribution over domains, i.e., a meta-distribution.
Specifically, we employ a novel distribution-level Universum strategy to generate supplementary diverse domain-related class-conditional distributions, thereby enhancing generalization.
This allows us to resample from these generated distributions to provide feedback to the primordial instance-level classifier, further improving its adaptability to the target-agnostic.
To ensure generation accuracy, we establish an additional distribution-level classifier to regularize these conditional distributions.
Extensive experiments have been conducted to demonstrate its effectiveness and low computational cost compared to SOTAs.
\end{abstract}


\section{Introduction}
\IEEEPARstart{I}{n} conventional machine learning, the independent identical distribution (i.i.d.) assumption \cite{vapnik1998statistical} posits that training and testing data are drawn from the same distribution.
However, real-world applications often violate this assumption due to variations in weather, environment, and other factors, a challenge known as domain shift \cite{nam2021reducing}.
For example, an autonomous driving model \cite{wang2020learning} trained in specific scenarios must adapt to diverse changes in weather, street conditions, and so on.
To handle domain shift and improve the generalizability for unseen scenarios, Domain Generalization (DG) has witnessed rapid development over the past decade \cite{wang2022generalizing}.

Various strategies have been proposed for DG, including data augmentation \cite{zhou2024mixstyle}, domain-invariant representation \cite{chevalley2022invariant}, meta-learning \cite{li2018learning}, etc.
Among them, domain-invariant representation (DIR) is a representative approach, which is fundamentally an solution for alignment distribution, seeking to achieve invariant representations across domains.
In other words, DIR expects that latent representations of each training domain belong to the same distribution, ensuring their applicability to arbitrary target domains.
Ganin et al. \cite{ganin2016domain} utilized adversarial learning to confuse the domain classifier to ensure alignment of the distributions.
Li et al. \cite{li2018domain} aligned infinite order moments of pairwise domains with kernel trick.
Salaudeen et al. \cite{salaudeen2024causally} extracted causal invariant representations from a causal perspective.

\begin{figure}[t]
	\centering
	\includegraphics[width=\linewidth]{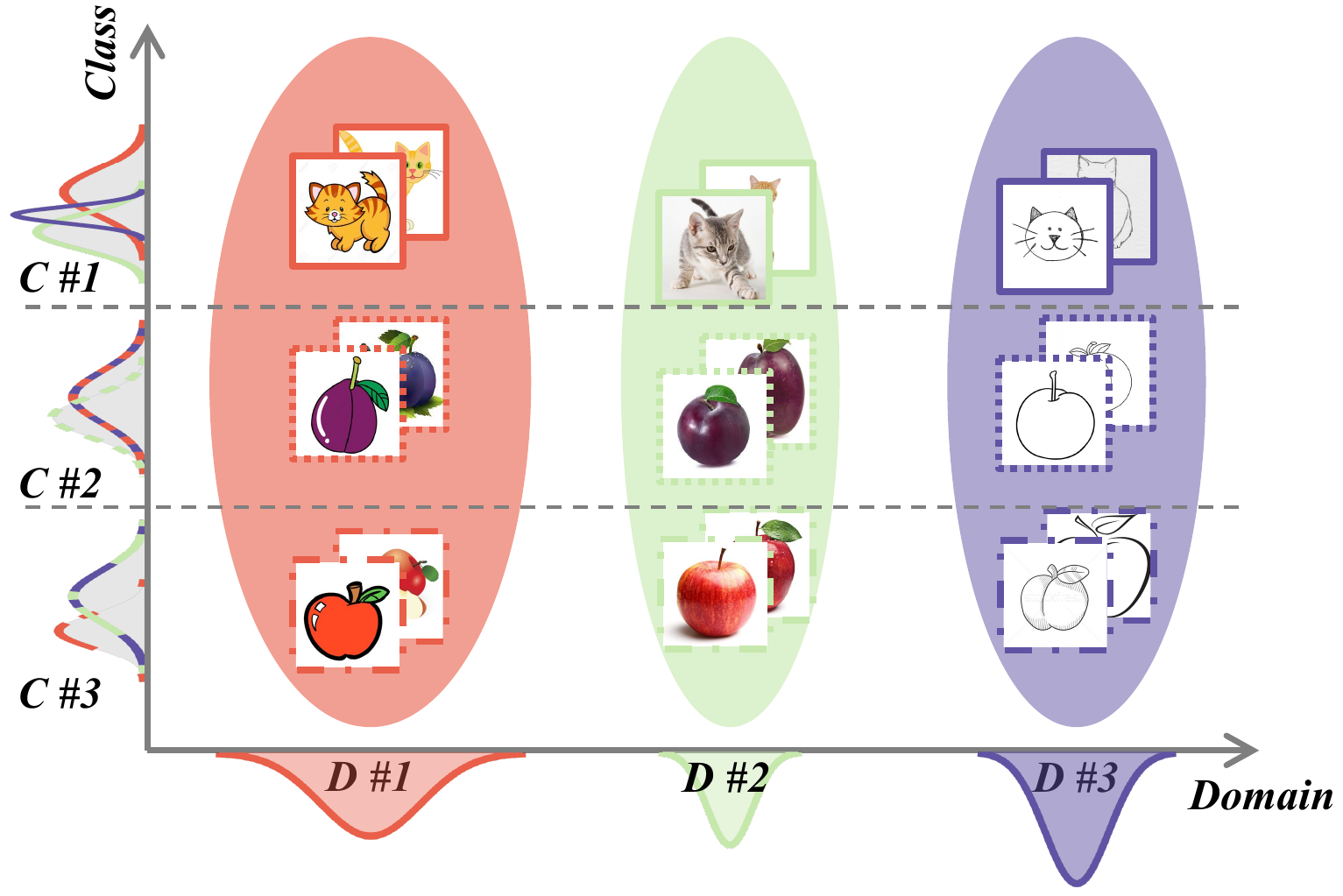}
	\caption{Illustration of two angles for DG. Different colors represent different domains, and different border lines represent different classes. From the domain angle, most existing approaches focus on their marginal distributions. From the class angle, multiple domain-related peaks have emerged within the same individual classes.}
	\label{fig:perspective}
\end{figure}

\begin{figure}[t]
	\centering 
	\includegraphics[width=\linewidth]{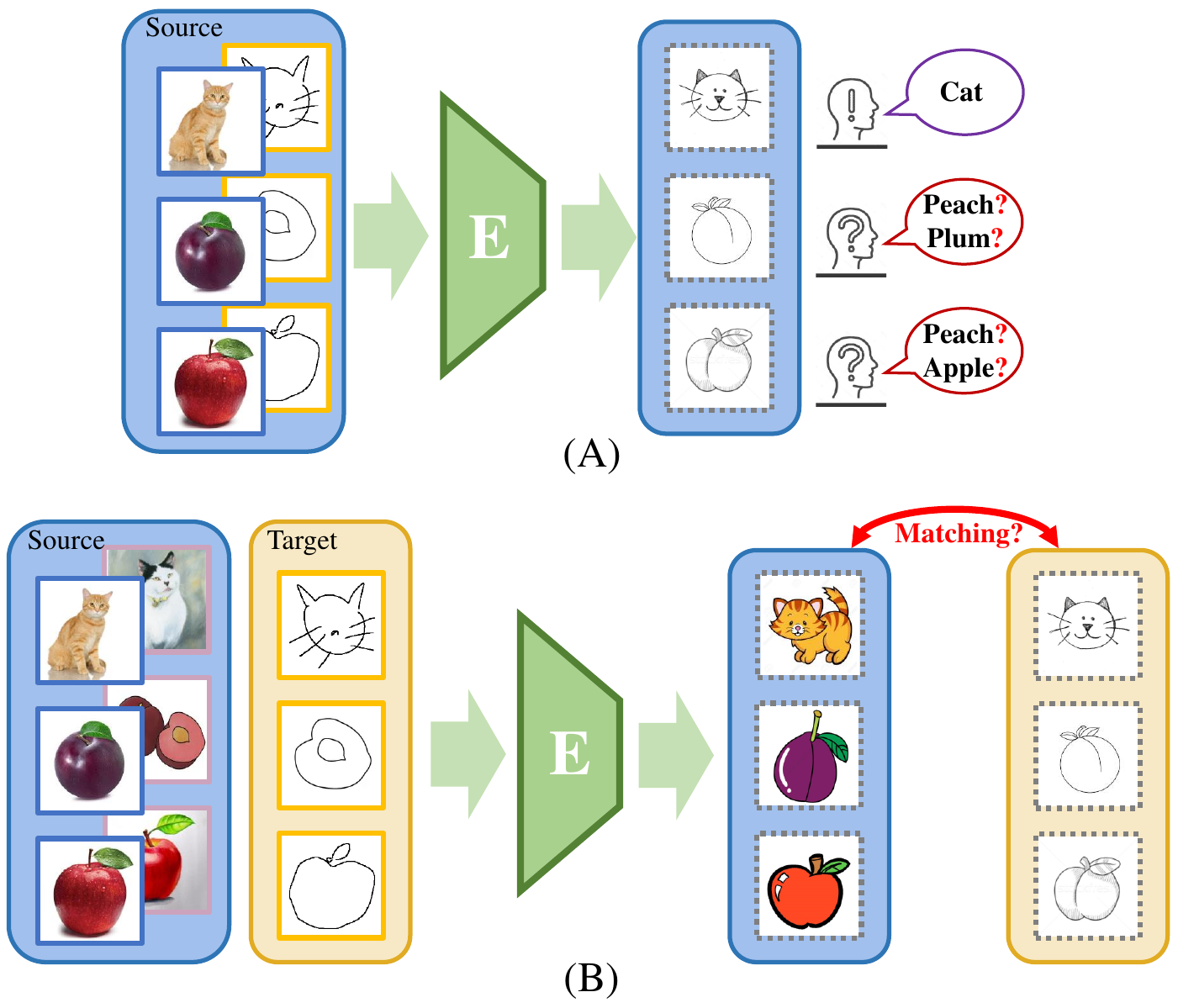}
	\caption{Illustration of two failed cases in DIR. The solid line squares denote the images sampled from the domains and the dashed squares denote the latent representations.}
	\label{fig:dir}
\end{figure}

Although DIR has yielded positive outcomes, it typically handles DG from \textit{a domain angle}, as illustrated in Fig. \ref{fig:perspective}, attempting to align their marginal distributions and leading to some practical challenges.
Consequently, in this paper, we regard DIR as an intuitive but ideal solution.
Intuitively, it is natural that when we collect adequate domains, DIR can effectively identify invariant representations across domains and achieve significant performance.
However, due to the high annotation cost, it may be difficult to meet the above case, which means that only a limited number of domains can be observed, leading to various issues that compromise DIR's effectiveness.
As shown in Fig. \ref{fig:dir}(A), in essence, DIR belongs to a subset of representations, which retain the least information compared to the information present in each domain \cite{xie2024enhancing}.
Therefore, DIR tends to discard the auxiliary yet simple discriminative information, complicating downstream tasks.
For example, apples and plums can be easily distinguished through color assistance, but when only in shape, they can be easily misjudged.
As shown in Fig. \ref{fig:dir}(B), when target domains are located outside the space spanned by observed domains, latent representations between the source and target domains may not align \cite{cao2024mixup}.
This mismatch highlights why Sketch as the target domain typically underperforms on PACS.
In summary, DIR (alignment distribution) may be unable to handle various unseen target domains.

\begin{figure*}[t!]
	\centering
	\includegraphics[width=\linewidth]{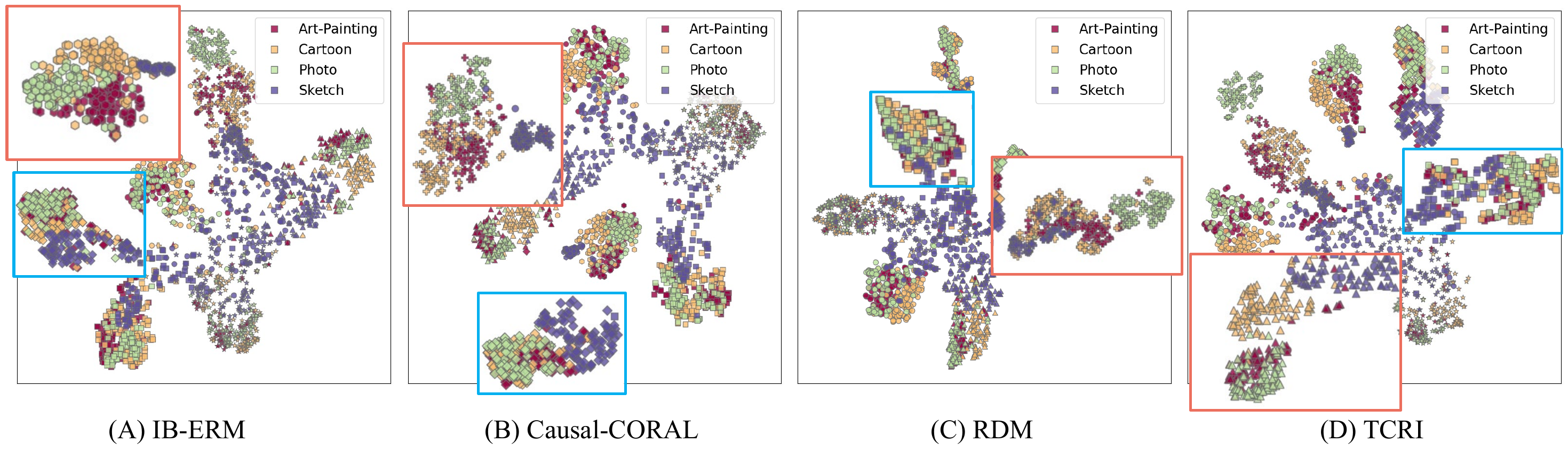}
	\caption{Visualization of polymorphic domain-related clusters within each individual classes testing in Sketch on PACS using t-SNE. Different colors correspond to different domains and different shapes correspond to different classes.}
	\label{fig:apx:clustering}
\end{figure*}

The above issues are discussed from the domain angle.
However, when switching to \textit{the class angle}, as illustrated by the vertical axis in Fig. \ref{fig:perspective}, more complicated problems have arisen.
Due to conditional distribution shift \cite{zhu2022localized}, polymorphic domain-related peaks or clusters may emerge within each class.
In particular, as shown in Fig. \ref{fig:perspective}, the number of clusters varies within each individual classes, with some being able to observe all domain clusters while others have partial overlap.
We have examined four recent DIR approaches and visualized them in Fig. \ref{fig:apx:clustering} using t-SNE in subsection \textit{Empirical Findings}, which consistently exhibit this phenomenon.
Incidentally, similar phenomena have also been observed by \cite{zhou2024mixstyle} and \cite{zhu2022localized}.
Therefore, in this case, it becomes more challenging to align their distributions.
One solution would be to add an explicit alignment for condition distributions, such as \cite{zhu2022localized, liya2018domain, wang2024enhancing}.
Nevertheless, the above issues illustrated in Fig. \ref{fig:dir} persist, or may even be exacerbated.
Therefore, we need to explore another solution for DG.

Let us once again focus on the class angle (axis) in Fig.\ref{fig:perspective}.
If we consider a domain-related cluster, i.e., $P^d\left(X \vert y\right)$, as a hyper-instance, a classification scenario at the distribution level could be constructed, analogous to a conventional instance classification scenario.
In other words, from the class angle, each domain-related cluster represents a distribution-level semantic point, and polymorphic domain-related clusters can be analogized to diverse samples in conventional scenario.
In this way, distribution alignment can be viewed as \textit{collapsing all hyper-instances} within each individual classes into \textit{a single hyper-instance}.
Therefore, alignment can be considered as an obstacle to representation, and its learned model will face significant challenges when handling various unseen target domains.
In contrast, in this paper, we aim to preserve these diverse hyper-instances to guide the learning of the model, as diversity can improve generalizability, just as in conventional learning.
Furthermore, the generalization bounds proposed in \cite{blanchard2021domain} and \cite{cao2024mixup}, invloved in the number of domains, provide theoretical assurance for this intuition.

To this end, we design a novel Conjugate Consistent Enhanced Module, namely \ours, based on a distribution over domains, i.e., a meta-distribution $\tau$.
The underlying intuition is that if the domain-related clusters aforementioned, denoted by $P^d\left(X \vert y\right)$, can be classified correctly, then the instances sampling from these distributions can also be naturally classified correctly.
Specifically, analogous to \cite{zhang2022class}, a novel distribution-level Universum strategy has been adopted to generate supplementary domain-related class-conditional distributions.
In this way, these generated distributions are diverse, scattered throughout the distribution space, and can indirectly better approximate the true meta-distribution $\tau$, thereby enhancing discriminative generalizability.
Notably, these generated distributions can also be regarded as simulations of the unseen targets.
Consequently, we can resample from these generated distributions to provide feedback to primordial instance classifier, thereby enhancing its adaptability to the target-agnostic.
To ensure generation accuracy, a distribution-level classifier should be established to regularize these class-conditional distributions.
Our main contributions are highlighted as follows:
\begin{itemize}
	\item To guide rather than obstruct, we propose a novel Conjugate Consistent Enhanced Module, namely \ours.
	\item A novel distribution-level Universum strategy is applied to generate additional conditional distributions to assist the primordial instance-level classifier.
	\item Compared to SOTAs, our \ours {} is simpler and can achieve lower computational cost.
	\item Extensive experiments have beens constructed to demonstrate the effectiveness of our \ours.
\end{itemize}

The rest of this paper is organized as follows.
Section \ref{sec:related} reviews related works.
Section \ref{sec:method} introduces proposed \ours {} in detail.
Section \ref{sec:exp} constructs extensive experiments to demonstrate its effectiveness and low computational cost compared to SOTAs.
Section \ref{sec:conclusion} draws some conclusions.

\section{Related Works} \label{sec:related}

This section will provide a detailed review of the two taxonomies utilized in DG: alignment and non-alignment.

The alignment distribution strategy is both intuitive and widely utilized, in which almost approaches can be summarized into domain-invariant representation (DIR).
To achieve invariant representations across domains, various approaches have been developed, which can be further concluded into three methodologies: kernel-based, adversarial-based, and causal-based approaches.
Li et al. \cite{li2018domain} leveraged the kernel mean embedding to align arbitrary order moments of two distributions.
Nguyen et al. \cite{nguyen2023domain} pointed out that the existing kernel-based approaches may suffer from the curse of dimensionality due to the sparsity of high-dimensional representation space.
Therefore, they turned to align risk distributions to indirectly achieve invariant representations.
Ganin et al. \cite{ganin2016domain} introduced the adversarial learning into DG to confuse the domain classifier to ensure alignment of the distributions.
Zhang et al. \cite{zhang2023free} pointed out that the distributions have disjoint support with high probability, leading to unstable of the domain classifier.
Therefore, they introduced a smoothing factor to alleviate this issue.
Li et al. \cite{li2022invariant} and Salaudeen et al. \cite{salaudeen2024causally} extracted causal invariant representations with the discrepant causal conditions.
The remaining approaches for alignment marginal distritbution are to match the gradient \cite{rame2022fishr, wang2023pgrad}, or to optimize the worst distributions followed by regrouping the samples \cite{sagawa2020distributionally}.
Meanwhile, there are some existing apporaches that attempt to match conditional distributions.
Li et al. \cite{liya2018domain} extracted class-oriented causal invariant represnetations.
Zhu et al. \cite{zhu2022localized} aligned localized class conditional distributions through adversarial learning.
Wang et al. \cite{wang2024enhancing} utilized two frozen orthogonal heads to extract invariant representations.
Nevertheless, the above issues illustrated in Fig. \ref{fig:dir} still persist and have not been resolved.
Therefore, our \ours {} does not follow the design concept of alignment distribution.

The non-alignment strategy can be analogous to conventional classification approaches.
Mixup \cite{yan2020improve} provided more interpolated instances to enhance the smoothness, thereby improving generalizability.
MLDG \cite{li2018learning} utilized the meta-learning to simulate the marginal shift between domains.
SIMPLE \cite{li2023simple} integrated the results of multiple checkpoints through ensemble learning, which provided diversity from the model aspect.
SAGM \cite{wang2023sharpness} sought a flat optimized region by minimizing a perturbed model.
Most approaches abovementioned focus on the shift in marginal distributions.
Nevertheless, we observe that mutliple domain-related clusters still emerge within each individual classes, even in the alignment approaches.
To this end, in analogy to conventional instance-level learning, we intuitively endeavor to leverage this phenomenon to generate diverse domain-related class-conditional distributions rather than to align them, where we indirectly generate adequate domains, to enhance generalizability.
Furthermore, Blanchard et al. \cite{blanchard2021domain} and Cao et al. \cite{cao2024mixup}, which proposed the generalization bounds involved in the number of domains, provided theoretical assurance for this intuition.

\section{Methodology} \label{sec:method}

In this section, a more detailed discussion will be presented.

\subsection{Preliminaries}

In DG, a domain can be defined as a joint distribution $P\left(X, Y\right)$ \cite{wang2022generalizing}, where $X$ denotes the instance space and $Y$ denotes the label space.
In this way, domain shift can be regarded as the changes between these joint distribution, i.e., $P^{d_i}\left(X, Y\right) \neq P^{d_j}\left(X, Y\right), \forall d_i \neq d_j$.
In this paper, $N$ domains can be observed, and they have the same label space, i.e., $\mathcal{Y}^{d_i} = \mathcal{Y}^{d_j}$.
Each domain can observe $n_i$ instance-label pairs, where $i \in \left\{1, \dots, N\right\}$.
$\tau$ denotes the true meta-distribution, i.e., a distribution over domains:
\begin{equation} \label{eq:px2pxy}
	\begin{aligned}
		{} & P^d\left(X, Y\right) \sim \tau, \textnormal{ and} \\
		P^d\left(X, Y\right) & = P\left(Y \vert X\right) \underbrace{P^d\left(X\right)}_{\textnormal{shift}} \\
		{} & = P\left(Y \vert X\right) \underbrace{ \int{P^d\left(X \vert y\right)} P\left(y\right) \textnormal{d}y }_{\textnormal{mariginal distribution}} \\
		{} & \approx P\left(Y \vert X\right) \sum_{y}{\underbrace{P^d\left(X \vert y\right)}_{\textnormal{conditional distirbution}} P\left(y\right)}
	\end{aligned}
\end{equation}
Here, each domain can be uniformly sampled from $\tau$, implying that $P\left(d\right)$ follows a uniform distribution, and the conditional distribution $P\left(Y \vert X\right)$ is invariant across domains, which is consistent with the general setting in DG.
Moreover, the prior distribution of the class $P\left(y\right)$ can also maintain consistency across domains.
From Eq. \eqref{eq:px2pxy}, it is evident that marginal $P^d\left(X\right)$ alignment does not necessarily guarantee conditional $P^d\left(X \vert y\right)$ alignment, leading to the occurrence of polymorphic domain-related peaks or clusters.

\subsection{Empirical Findings}

In this subsection, we will describe the phenomenon of polymorphic domain-related class clusters in detail.

Fig. \ref{fig:apx:clustering} has visualized the latent representations extracted through four current representative domain-invariant representation approaches.
Therefrom we can find that 1) domain-related class clusters must emerge, and 2) these clusters are polymorphic.
Intuitively, these phenomena should not occur, instead the invariant representations should not be able to distinguish domain labels.
In other words, even with the alignment distribution approaches, the conditional distributions are still not perfectly aligned.
RDM in Fig. \ref{fig:apx:clustering}(C) seems to be the best effective approach, where half of the classes of the source domains appear to be aligned.
However, it still retains unaligned classes, not to mention that the target domain is not close to any source domain.
One possibility is that the approach itself can not align the observed domains.
Nevertheless, diverse polymorphism within involved approaches can refute such possibility, where within each individual classes, some can observe all domain clusters while others have partial overlap, illustrated in the rectangular box in Fig. \ref{fig:apx:clustering}.
This fact further indicates that various alignment distribution approaches have tried their best to align, but their effectiveness is limited due to various class-conditional distributions.
Consequently, we argue that marginal alignment does not necessarily guarantee conditional alignment.
Meanwhile, the target domain typically underperforms with these alignment dstributions, which are not aligned.
To this end, we seek a way to explore and exploit these domain-related clusters to improve generalizability. 

\subsection{Theoretical Evidence}

\begin{figure*}[t]
	\centering
	\includegraphics[width=\linewidth]{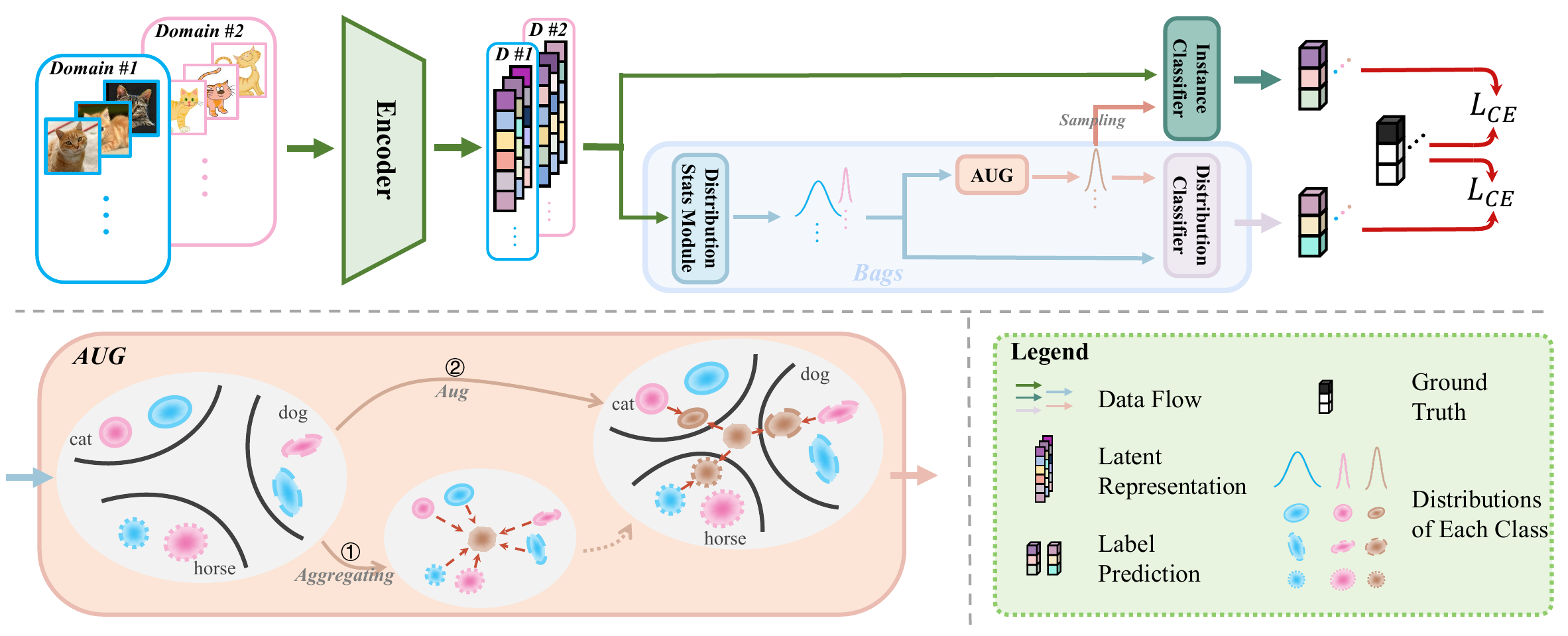}
	\caption{Illustration of a pipeline during training. Our \ours {} can be regarded as an additional module, which contains three components: 1) Distribution Statistics Module, 2) Augmentation Manipulation (AUG), and 3) Distribution-Level Classifier.}
	\label{fig:framework}
\end{figure*}

\begin{theorem}
	{\textnormal{\bfseries (Generalization Bound for DG based on PAC-Bayesian framework \cite{cao2024mixup})}}
	Giving a hypothesis space $\mathcal{H}$, and  $N$ domains $\left\{D_n\right\}^N_{n=1}$ sampled from $\tau$, where each domain $D_n$ consists of $M_n$ samples.
	Let $\mathcal{P}$ denotes a hyper-prior distribution  $\mathcal{P} \in \mathcal{M}\left(P\right)$, where $P \in \mathcal{M}\left(\mathcal{H}\right)$ and $\mathcal{M}\left( \mathcal{S} \right)$ denotes the set of all probability over $\mathcal{S}$.
	Then, for any $\delta \in \left(0, 1\right]$, the following inequality holds uniformly for all hyper-posterior distributions $\mathcal{Q}$ with probablity at least $1-\delta$:
	\begin{equation*}
		\begin{aligned}
			\mathbb{E}_{Q \sim \mathcal{Q}}{er\left(Q, \tau\right)} \leq & \frac{1}{N}\sum\limits_{n=1}^{N}{\mathbb{E}_{P \sim \mathcal{Q}}{\hat{er}\left(P, D_n\right)}} \\		
			{} & + \sqrt{\frac{L_0 \cdot W_1\left(\mathcal{Q}, \mathcal{P}\right) + \ln \left(N / \delta\right)}{2\left(N - 1\right)}} \\
			{} & + \frac{1}{N}\sum\limits_{n=1}^{N}{\sqrt{\frac{L_n \cdot W_1\left(Q, P_n\right) + \ln \left(NM_n / \delta\right)}{2\left(M_n - 1\right)}}}
		\end{aligned}
	\end{equation*}
	where $W_1\left(\cdot, \cdot\right)$ is the 1st order Wasserstein Distance, and $L_0$ and $L_n$ are the Lipschitz constants.
\end{theorem}

We can easily discover that the generalization error of the target domain is inversely proportional to the number of the observed domains or the source doamins, i.e., $N$.
Therefore, when the source domains are more diverse, the generalization error will be lower.
Our \ours {} aims to generate sufficient domain-related class-conditional distributions, i.e., $P^d\left(X \vert y\right)$.
Accoding to Eq. \eqref{eq:px2pxy}, our \ours {} can indirectly generate a large number of pseudo domains $P$, consisted of these $P^d\left(X \vert y\right)$.
In this way, we can conclude that the meta-distribution $\tau$ can be better approximated through our \ours, thereby enhancing generalizability.

\subsection{\ours}

Our \ours {} has been visualized in detail in Fig. \ref{fig:framework}, which is interfaced after the representation layer and parallelized with the primordial instance-level classifier.

\vspace{0.45em}
\noindent \textbf{Distribution Statistics Module} This component consists of two Full Connected layers to capture first-order and second-order statistics, namely mean and variance, respectively.
In Bag learning \cite{zhang2022multi}, a series of instances can be viewed as a whole, termed a ``bag", which can be represented as a distribution.
Inspired by this concept, we herein consider instances from the same class and domain as a bag, which is the domain-related cluster aforementioned.
Following the widely-used statistical method in \cite{kingma2013auto, zhang2022multi}, each mean $\mu_k^d$ and each standard deviation $\sigma_k^d$ can be formulated:
\begin{gather}
	\mu_i^d = FC_{\mu}\left(z_i^d\right), \sigma_i^d = \exp(0.5 * FC_{logvar}\left(z_i^d\right)) \label{eq:stats} \\
	\mu_k^d = Agg\left(\left\{\mu_i^d \vert y_i = k\right\}\right), \sigma_k^d = Agg\left(\left\{\sigma_i^d \vert y_i = k\right\}\right) \label{eq:stats:agg}
\end{gather}
\noindent where $z_i^d$ denotes the latent representation of $i$-th instance sampling from $d$-th domain.
In this way, $u_i^d$ and $\sigma_i^d$ denote the corresponding mean and standard deviation, and $FC_{\mu}$ and $FC_{logvar}$ denote the corresponding FC layer, respectively.
$k$ denotes $k$-th class, and $Agg\left(\cdot\right)$ denotes an aggregation method.
In this paper, average aggregation is adopted.

\vspace{0.45em}
\noindent \textbf{Distribution-Level Classifier} To protect semantic information of $\mu_k^d$ and $\sigma_k^d$, a distribution-level classifier should be established to discriminate each class.
Furthermore, the effectiveness of generated supplementary class distributions or domain-related clusters can be guaranteed by the foundation.
According to a distribution-free distribution regression approach proposed in \cite{poczos2013distribution}, we can design a distribution classifier with the kernel trick.
Therefore, the formula of the kernel embedding for a certain distribution $p$ can be represented as follows:
\begin{equation} \label{eq:embedding}
	e_p = K\left(p; \left\{P_k^d \vert k \in Y, d \in 1, \cdots, N \right\}\right) \in \mathbb{R}^{1 \times N\vert Y \vert}
\end{equation}
\noindent And, each element can be formulated as:
\begin{equation}
	\left(e_p\right)_k^d = K\left(p; P_k^d\right) = \sum_{h_i}{\exp\left(-\frac{D\left(p, P_k^d\right)}{h_i}\right)}
\end{equation}
\noindent where $K\left(\cdot; \cdot\right)$ denotes a distribution-level kernel function, in which each input is a distribution.
Here, the commonly used Radial Basis Function (RBF) kernel \cite{li2018domain} is adopted.
Along this line, $D\left(\cdot, \cdot\right)$ denotes the difference between two distributions, and $h_i$ denotes the radial in RBF.
Following the implementation in \cite{li2018domain} and \cite{salaudeen2024causally}, multiple $h_i$ are also adopted to avoid spending time choosing the appropriate $h_i$, thereby reducing the time complexity.

Subsequently, we can feed acheived embedding $e_p$ by Eq. \eqref{eq:embedding} into a projection head $H\left(\cdot\right)$ for classification, similar to the instance classifier.
Therefore, the prediction of $p$ in our \ours {} can be represented as:
\begin{equation} \label{eq:ours:pred}
	\widehat{pred}_p = H\left(e_p\right)
\end{equation}

To measure the difference between two distributions, i.e., $D\left(p, P_k^d\right)$, 2-Wasserstein distribution \cite{panaretos2019statistical} under Gaussian Distribution is adopted:
\begin{equation}
	\begin{aligned}
		D^2\left(p, P_k^d\right) & = D^2\left(\left(\mu, \sigma\right), \left(\mu_k^d, \sigma_k^d\right)\right) \\
		{} & = \lVert \mu - \mu_k^d \rVert_2^2 + \lVert \sigma - \sigma_k^d \rVert_2^2
	\end{aligned}
\end{equation}

During training, in each batch, complete distribution information about $P_k^d$ can not be acheived.
As an alternative, we adopt corresponding historical distribution information $\left(P_h\right)_k^d$ with a momentum strategy, and its updating formula can be represented as follows:
\begin{equation} \label{eq:p:hist}
	\left(P_h\right)_k^d = \rho \left(P_{h-1}\right)_k^d + \left(1 - \rho\right) P_k^d
\end{equation}
\noindent where $\rho$ denotes the momentum weight, which is often greater than 0.9, and $\left(P_{h-1}\right)_k^d$ denotes the last information.

\vspace{0.45em}
\noindent \textbf{Augmentation Manipulation} To better approximate the true hyper-distribution $\tau$, more conditional distributions or domain-related clusters are needed.
In this way, the discriminant surface can be better inferred to improve its generalizability, which can be theoretically guaranteed accroding to the generalization bounds proposed in \cite{blanchard2021domain} and \cite{cao2024mixup}.

Similar to conventional classification scenario, Mixup \cite{yan2020improve} strategy can be a solution to generate these conditional distributions.
Nevertheless, in essence, Mixup is subject to the linear interpolation approach.
Here, on the one hand, the generated instances by mixing instance pairs from the same class still locate inside the space spanned by observed instances.
This case cannot address the shortcoming raised in Fig. \ref{fig:dir}(B), which goes against our motivation.
On the other hand, the instances generated by mixing pairs of instances from the different classes are uncontrolled, which may be uncertain whether they are located in low-density areas \cite{you2019universal}.
Learnable optimization-based strategy, e.g., \cite{qiao2020learning}, seems to be another feasible solution.
Although it can overcome the shortcomings, it is too complex to employ.
We are seeking a solution as convenient as Mixup that generates instances outside the interpolation space while preserving semantic information.

Intuitively, directly mixing points from low-density areas with a particular semantic point appears to achieve the desired goal.
The reason behind it is that points from these areas often confuse the classifier to make decisions.
Consequently, the mixed points has a clear semantic bias.
In this paper, we employ Universum \cite{liu2024timesurl}, which represents a set of instances that do not belong to each observed class in conventional classification.
Unlike \cite{zhang2022class}, we should construct Universum in the distribution space.
In other words, our distribution-level Universum does not belong to the observed class conditional distributions.
Thanks to the view that domain-related clusters are regarded as hyper-instances, analogous to the traditional Universum, it is possible to construct its formula with relative ease.
As shown in step \textcircled{1} of AUG module in Fig. \ref{fig:framework}, we can aggregate all observed conditional distributions, i.e., $P\left(X \vert y\right)$, to obtain our distribution-level Universum:
\begin{equation} \label{eq:universum}
	P_u = Agg\left(\left\{P_k^d \vert k \in Y, d \in 1, \cdots, N\right\}\right)
\end{equation}
\noindent where $Agg\left(\cdot\right)$ also denotes the average aggregation.

To this end, in each batch, the generated supplementary class conditional distributions can be formulated as follows:
\begin{equation} \label{eq:pu}
	\left(P_u^{aug}\right)_k^d = (1 - \lambda) P_u + \lambda P_k^d
\end{equation}
\noindent where $\lambda$ denotes the mixing coefficient, which is fixed at 0.5.
It is worth noting that the corresponding class label is $k$.

These generated distributions serve two purposes.
One is to resample the instances to provide feedback to the primordial instance-level classifier for better generalizability.
The intuition behind it is that \textit{if the distribution-level classification is accurate, the corresponding sample level classification should also be correct.}
These instances are sampled from Gaussian Distribution, the same as VAE \cite{kingma2013auto}, formulated in Eq. \eqref{eq:resample}.
Another is to promote the distribution-level classifier learning.
\begin{equation} \label{eq:resample}
	\begin{aligned}
		& \left(x_u^{aug}\right)_k^d \sim \left(P_u^{aug}\right)_k^d \\
		\Rightarrow & \left(x_u^{aug}\right)_k^d \sim \mathcal{N}\left(\left(\mu_u^{aug}\right)_k^d, \left(\sigma_u^{aug}\right)_k^d\right)
	\end{aligned}
\end{equation}

\subsection{Total Loss}

For convenience, we take ERM as the foundation, illustrated in Fig. \ref{fig:framework}.
Given the restricted number of observed domains, the label smoothing strategy may be employed to enhance performance \cite{yan2020improve}.
To this end, Mixup has been implemented in the distribution space, where it is exclusively employed for distribution-level classification.
The formulation can be represented as follows:
\begin{equation} \label{eq:mixup}
	\begin{aligned}
		\hat{P}^{aug} & = \gamma P_{k_m}^{d_i} + \left(1 - \gamma\right) P_{k_n}^{d_j} \\
		\hat{y}^{aug} & = \gamma k_m + \left(1 - \gamma\right) k_n
	\end{aligned}
\end{equation}

\vspace{0.45em}
\noindent\textbf{Instance-Level Loss}
This loss contains two parts: 1) the cross-entropy loss induced by the instances sampling from the observed domains; 2) the cross-entropy loss induced by the instanced resampling from the generated distributoins.
\begin{equation} \label{eq:ins:loss}
	\begin{aligned}
		\mathcal{L}_{ins} = & \frac{1}{\sum_{d=1}^{N}n_d}\sum_{i=1}^N{\ell_{CE}\left(x_i^d, y_i\right)} \\
		{} & + \frac{1}{\sum_{d=1}^{N}n_d}\sum_{d=1}^N{\sum_{k=1}^{\lvert Y \rvert} \sum_i \ell_{CE}\left(\left[\left(x_u^{aug}\right)_k^d\right]_i, k\right)}
	\end{aligned}
\end{equation}
\noindent where $\ell_{CE}\left(\cdot, \cdot\right)$ denotes the cross-entropy loss.

\vspace{0.45em}
\noindent\textbf{Distribution-Level Loss}
This loss contains three parts: 1) the cross-entropy loss induced by the class conditional distributions from the observed domains; 2) the cross-entropy loss induced by the generated distributions; 3) the smoothness loss induced by mixup.
\begin{equation} \label{eq:dis:loss}
	\begin{aligned}
		\mathcal{L}_{dis} = & \frac{1}{N\lvert Y \rvert} \sum_{d=1}^{N}{\sum_{k=1}^{\lvert Y \rvert}{\ell_{CE}'\left(\widehat{pred}_k^d, k\right)}} \\
		{} & + \frac{1}{N\lvert Y \rvert} \sum_{d=1}^{N}{\sum_{k=1}^{\lvert Y \rvert}{\ell_{CE}'\left(\left(\widehat{pred}_u^{aug}\right)_k^d, k\right)}}\\
		{} & + \frac{1}{N\lvert Y \rvert} \sum_{d=1}^{N}{\sum_{k=1}^{\lvert Y \rvert}{\ell_{CE}'\left(\left(\widehat{pred}^{aug}\right)_k^d, \hat{y}\right)}}
	\end{aligned}
\end{equation}
\noindent where $\ell_{CE}'\left(\cdot, \cdot\right)$ denotes the cross-entropy loss, $\widehat{pred}$ can be calculated by Eq. \eqref{eq:ours:pred}, and $\hat{y}$ can be generated by Eq. \eqref{eq:mixup}. 

To sum up, the total objective function can be formulated:
\begin{equation} \label{eq:total}
	\mathcal{L} = \mathcal{L}_{ins} + \beta \mathcal{L}_{dis}
\end{equation}
\noindent where $\beta$ denotes the trade-off.
Algorithm \ref{algo:ours} summarizes the overall process of our proposed \ours.

\begin{algorithm}[t]
	\caption{\ours}
	\label{algo:ours}
	
	\textbf{Input:} Training set, batch size $B$, number of source domains $N$, number of class $\lvert Y \rvert$, maximum iteration $T$, Beta Distribution parameter $\alpha$, momentum weight $\rho$, the mixing coefficient $\lambda$, the trade-off $\beta$, the radials $h_i$
	
	\textbf{Parameters:} An encoder $E$, a instance-level classifier $\text{CLS}$, a mean statistics layer $\text{FC}_{\mu}$, a standard deviation statistics layer $\text{FC}_{\sigma}$, a distribution-level classifier $H$
	
	\textbf{Output:} Encoder $E$, instance-level classifier $\text{CLS}$
	
	\begin{algorithmic}[1]
		\FOR{$t \leftarrow 1$ to $T$}
		\STATE Randomly sample a batch $S = \left\{\left(x_b, y_b\right)\right\}_{b=1}^{B}$
		\\ \textit{\textcolor{gray}{\# Extract the latent representation}}
		\STATE $z_i \leftarrow E\left(x_i\right)$
		\\ \textit{\textcolor{gray}{\# Achieve the distribution statistics $\left(P_i = \left(\mu_i, \sigma_i\right)\right)$}}
		\STATE $P_i \leftarrow$ Eq. \eqref{eq:stats} with current batch
		\STATE $P_k^d \leftarrow$ Eq. \eqref{eq:stats:agg} with $\left\{\left(P_i, y_i\right)\right\}$
		\\ \textit{\textcolor{gray}{\# Update historical information}}
		\STATE $\left(P_h\right)_k^d \leftarrow$ Eq. \eqref{eq:p:hist} with $\rho$
		\\ \textit{\textcolor{gray}{\# Augmentation Manipulation}}
		\STATE $P_u \leftarrow$ Eq. \eqref{eq:universum} with $\left\{P_k^d\right\}$
		\STATE $\left(P_u^{aug}\right) \leftarrow$ Eq. \eqref{eq:pu} with $\lambda$
		\\ \textit{\textcolor{gray}{\# Smoothing distribution-level classifier}}
		\STATE $\left(\hat{P}^{aug}, \hat{y}^{aug}\right) \leftarrow$ Eq. \eqref{eq:mixup}
		\STATE $\mathcal{L}_{dis} \leftarrow$ Eq. \eqref{eq:dis:loss} with $H$
		\\ \textit{\textcolor{gray}{\# Resample the instances}}
		\STATE $\left(x_u^{aug}\right)_k^d \leftarrow$ Eq. \eqref{eq:mixup} with $\gamma \sim \text{Beta}\left(\alpha, \alpha\right)$
		\STATE $\mathcal{L}_{ins} \leftarrow$ Eq. \eqref{eq:ins:loss} with $\text{CLS}$
		\STATE $\mathcal{L} \leftarrow$ Eq. \eqref{eq:total} with $\beta$
		\STATE Update network parameters
		\ENDFOR
	\end{algorithmic}
\end{algorithm}

\section{Experiments} \label{sec:exp}

In this section, extensive experiments have been constructed to comprehensively evaluate the effectiveness of \ours {} on six Benchmarks.

\subsection{Datasets and Settings}

For the architecture, our \ours {} aligns with the settings outlined in DomainBed \cite{gulrajani2021in}, wherein ResNet-50 has been selected as the backbone and the training-domain validation method has been employed for model selection.
Extensive experiments are constructed on six Benchmarks, i.e., C-MNIST (ColoredMNIST), VLCS, PACS, OfficeHome, TerraIncognita, and DomainNet.
The average results on these dataset are reported in Tab. \ref{tab:result_avg}.

Eighteen recent strong comparison approaches and two other representative approaches are deplyed to compare with our \ours.
These approaches can be divided into six categories: 1) \textit{distribution robust method} (GroupDRO \cite{sagawa2020distributionally}), 2) \textit{causal methods learning invariance} (IRM \cite{arjovsky2019invariant}, VREx \cite{krueger2021out}, EQRM \cite{eastwood2022probable}, TCRI \cite{salaudeen2024causally}), 3) \textit{gradient matching} (Fish \cite{shi2022gradient}, Fishr \cite{rame2022fishr}, SAGM \cite{wang2023sharpness}, PDGrad \cite{wang2023pgrad}), 4) \textit{representation distribution matching} (MMD \cite{li2018domain}, CORAL \cite{sun2016deep}, RDM \cite{nguyen2023domain}), 5) \textit{meta-learning} (MLDG \cite{li2018learning}, ARM \cite{zhang2021adaptive}) 6) and \textit{other variants} (Mixup \cite{yan2020improve}, SagNet \cite{nam2021reducing}, Mixstyle \cite{zhou2024mixstyle}).

\begin{table*}[t]
	\centering
	\caption{Overall accuracy results on five Benchmarks. The \textbf{bold}, \underline{underline}, and \textit{italic} items are the best, the second-best, and the third-best results, respectively. $\dagger$ denotes our reproduced results.}
	\label{tab:result_avg}
	\begin{tabular}{l ccccccc}
		\toprule
		Methods & C-MNIST & VLCS  & PACS & OfficeHome & TerraIncognita & DomainNet & Avg \\
		\midrule
		ERM \citeay{vapnik1998statistical}{1998} & 51.5 $\pm$ 0.1 & 77.5 $\pm$ 0.4  & 85.5 $\pm$ 0.2  & 66.5 $\pm$ 0.3 & 46.1 $\pm$ 1.8 & 40.9 $\pm$ 0.3 & 61.3 \\
		Mixup \citeay{yan2020improve}{2020} & \textit{52.1 $\pm$ 0.2} & 77.4 $\pm$ 0.6  & 84.6 $\pm$ 0.6 & 68.1 $\pm$ 0.3 & 47.9 $\pm$ 0.8 & 39.2 $\pm$ 0.4 & 61.5 \\
		DANN \citeay{ganin2016domain}{2016} & 51.5 $\pm$ 0.3 & 78.6 $\pm$ 0.7  & 83.6 $\pm$ 1.1 & 65.9 $\pm$ 0.6 & 46.7 $\pm$ 1.6 & 38.3 $\pm$ 0.3 & 60.8 \\
		MLDG \citeay{li2018learning}{2018} & 51.5 $\pm$ 0.2 & 77.2 $\pm$ 0.4  & 84.9 $\pm$ 1.0 & 66.8 $\pm$ 0.6 & 47.7 $\pm$ 0.9 & 41.2 $\pm$ 0.3 & 61.5 \\
		GroupDRO \citeay{sagawa2020distributionally}{2020} & \textit{52.1 $\pm$ 0.2} & 76.7 $\pm$ 0.6  & 84.4 $\pm$ 0.8 & 66.0 $\pm$ 0.7 & 43.2 $\pm$ 1.1 & 33.3 $\pm$ 0.2 & 59.3 \\
		IRM \citeay{arjovsky2019invariant}{2019} & 52.0 $\pm$ 0.3 & 78.5 $\pm$ 0.5  & 83.5 $\pm$ 0.8 & 64.3 $\pm$ 2.2& 47.6 $\pm$ 0.8 & 33.9 $\pm$ 1.9 & 60.0 \\
		ARM \citeay{zhang2021adaptive}{2021} & 51.9 $\pm$ 0.5 & 77.6 $\pm$ 0.7 & 85.1 $\pm$ 0.7 & 64.8 $\pm$ 0.4 & 45.5 $\pm$ 1.3 & 35.5 $\pm$ 0.5 & 60.1 \\
		VREx \citeay{krueger2021out}{2021} & 51.8 $\pm$ 0.2 & 78.3 $\pm$ 0.2  & 84.9 $\pm$ 0.6 & 66.4 $\pm$ 0.6 & 46.4 $\pm$ 0.6 & 33.6 $\pm$ 3.0 & 60.2 \\
		EQRM \citeay{eastwood2022probable}{2022} & 52.0 $\pm$ 0.2 & 77.8 $\pm$ 0.6  & 86.5 $\pm$ 0.2 & 67.5 $\pm$ 0.3 & 47.8 $\pm$ 0.6 & 41.0 $\pm$ 0.6 & 62.1 \\
		Fish \citeay{shi2022gradient}{2022} & 51.4 $\pm$ 0.1 & 77.8 $\pm$ 0.3  & 85.5 $\pm$ 0.3 & 68.6 $\pm$ 0.4 & 45.1 $\pm$ 1.3 & 42.7 $\pm$ 0.2 & 61.8 \\
		Fishr \citeay{rame2022fishr}{2022} & 51.8 $\pm$ 0.1 & 77.8 $\pm$ 0.1  & 85.5 $\pm$ 0.4 & 67.8 $\pm$ 0.1 & 47.4 $\pm$ 1.6 & 41.7 $\pm$ 0.3 & 62.0 \\
		CORAL \citeay{sun2016deep}{2016} & 51.5 $\pm$ 0.2 & 78.8 $\pm$ 0.6  & 86.2 $\pm$ 0.3 & 68.7 $\pm$ 0.3 & 47.6 $\pm$ 1.0 & 41.5 $\pm$ 0.2 & 62.4 \\
		MMD \citeay{li2018domain}{2018} & 51.5 $\pm$ 0.3 & 77.5 $\pm$ 0.9  & 84.6 $\pm$ 0.5 & 66.3 $\pm$ 0.1 & 42.2 $\pm$ 1.6 & 23.4 $\pm$ 9.5 & 57.6 \\
		SagNet \citeay{nam2021reducing}{2021} & 51.7 $\pm$ 0.1 & 77.8 $\pm$ 0.7 & 86.3 $\pm$ 0.5 & 68.1 $\pm$ 0.3 & 48.6 $\pm$ 1.8 & 40.3 $\pm$ 0.3 & 62.1 \\
		Mixstyle \citeay{zhou2024mixstyle}{2024} & 52.0 $\pm$ 0.2 & 77.6 $\pm$ 0.9 & 85.2 $\pm$ 1.1 & 60.4 $\pm$ 0.5 & 44.0 $\pm$ 1.3 & 34.0 $\pm$ 0.3 & 58.9 \\
		RDM \citeay{nguyen2023domain}{2023} & \underline{52.2 $\pm$ 0.2} & 78.4 $\pm$ 0.4  & \textit{87.2 $\pm$ 0.7} & 67.3 $\pm$ 0.4 & 47.5 $\pm$ 1.0 & 43.4 $\pm$ 0.4 & 62.7 \\
		SAGM \citeay{wang2023sharpness}{2023} & 51.7 $\pm$ 0.3 & \textit{80.0 $\pm$ 0.4} & 86.6 $\pm$ 0.3 & \textbf{70.1 $\pm$ 0.2} & 48.8 $\pm$ 0.8 & \underline{45.0 $\pm$ 0.2} & \textit{63.7} \\
		PGrad \citeay{wang2023pgrad}{2023} & 51.9 $\pm$ 0.2 & 78.9 $\pm$ 0.5 & 86.2 $\pm$ 0.6 & \textit{69.8 $\pm$ 0.3} & 50.7 $\pm$ 1.0 & 41.0 $\pm$ 0.2 & 63.1 \\
		TCRI$\dagger$ \citeay{salaudeen2024causally}{2024} & \textit{52.1 $\pm$ 0.3} & 79.9 $\pm$ 0.6 & 87.0 $\pm$ 0.5 & 67.9 $\pm$ 0.4 & \textit{51.2 $\pm$ 0.8} & 42.2 $\pm$ 0.1 & 63.4 \\
		\midrule
		\ours~(\textit{ours}) & \textbf{52.3 $\pm$ 0.3} & \underline{80.2 $\pm$ 0.5} & \underline{87.3 $\pm$ 0.6} & 67.5 $\pm$ 0.5 & \underline{52.1 $\pm$ 1.4} & \textit{44.8 $\pm$ 0.3} & \underline{64.0} \\
		\ours-L~(\textit{ours}) & \textbf{52.3 $\pm$ 0.3} & \textbf{80.3 $\pm$ 0.4} & \textbf{87.6 $\pm$ 0.5} & \underline{70.0 $\pm$ 0.3} & \textbf{52.2 $\pm$ 1.2} & \textbf{45.2 $\pm$ 0.2} & \textbf{64.6} \\
		\bottomrule
	\end{tabular}
\end{table*}

\begin{table}[t]
	\centering
	\caption{The settings of specific parameters involved on PACS. $-$ denotes a Fixed parameter without random range. The difference between \ours {} and \ours-L is only in the batch size setting, where \ours-L follows the setting in RDM \cite{nguyen2023domain}.}
	\label{tab:settings}
	\scriptsize
	\tabcolsep=0.62em
	\begin{tabular}{l | c c | c c}
		\toprule
		& \multicolumn{2}{c|}{\ours} & \multicolumn{2}{c}{\ours-L} \\
		\cmidrule{2-5}
		& Default & Random & Default & Random \\
		\midrule
		Learning Rate & 5e-5 & $10^{[-5, -3.5]}$ & 5e-5 & $10^{[-5, -3.5]}$ \\
		Batch Size & \textcolor{blue}{32} & \textcolor{blue}{$10^{[3, 5.5]}$} & \textcolor{red}{88} & \textcolor{red}{[70, 100]}\\
		Beta Distribution Parameter $\alpha$ & 0.2 & $10^{[-3, -1]}$ & 0.2 & $10^{[-3, -1]}$ \\
		Trade-off $\beta$ & 1e-2 & $10^{[-1, 1]}$ & 1e-2 & $10^{[-1, 1]}$ \\
		Momentum Weight $\rho$ & 0.95 & $-$ & 0.95 & $-$ \\
		Mixing Coefficient $\lambda$ & 0.5 & $-$ & 0.5 & $-$ \\
		\bottomrule
	\end{tabular}
\end{table}

For detailed implementation, each input image is resized and cropped to 224$\times$224.
Adam is utilized as the optimizer.
The training data are randomly split into two parts: 80\% for training and 20\% for validation.
The best model on the validation split is selected to evaluate the target domain.
Maximum iteration $T$ is set to 5,000.
Taking PACS as an example, the remaining specific parameter settings involved are reported in Tab. \ref{tab:settings}.
The codes are run on Python 3.9, PyTorch 1.13 on Arch Linux with an NIVIDIA GeForce RTX 4090 GPU.

\subsection{Results and Analysis}

From Tab. \ref{tab:result_avg}, we can observe the following findings:
\textit{1)} Overall, our \ours {} can achieve almost the best results both on each dataset and average.
In particular, compared to the SOTA, our \ours {} gains the performances with the improvement of 0.9$\%$ on TerraIncognita, and 0.3$\%$ on average, which are significant.
Additionally, our \ours {} is simpler than the SOTA, particularly compared to SAGM and PGrad.
These observations can demonstrate the effectiveness of our proposed \ours.
\textit{2)} The result of \ours {} on OfficeHome seems to be less than satisfactory.
The reason behind it may be that the number of classes is typically greater than the batch size, resulting in insufficient sampling for each class.
In other words, domain-related class-conditional distributions are estimated with a greater bias on OfficeHome.
Nevertheless, compared to the baseline, i.e., ERM, our \ours {} still achieves the improvements.
Consequently, it means that \ours {} can be effective.
Herein, one possible solution to boost performance of \ours {} is to increase batch size.
For fairness, however, we still reported this result without any change to the batch size setting implemented in DomainBed.
\textit{3)} Combined with Tab. \ref{tab:app_pacs} in the Appendix, our \ours {} achieve the significant improvement in Sketch domain on PACS.
This domain carries less information compared to the source domains, and can cause potential mismatches in latent representations with DIR approaches.
Therefore, compared to the vast majority of DIR approaches, our \ours {} performs better and is more stable than RDM.
Moreover, \ours {} has achieves the greatest improvement in Cartoon Domain on PACS, which is another target domain with difficulty in generalization.
These observations can demonstrate that due to the non-alignment, \ours {} can escape from the two issues mentioned in Fig. \ref{fig:dir} and achieve more generalizability.
\textit{4)} Our \ours {} has achieved better performanance than other data augmentation approaches, e.g., Mixup and Mixstyle, on all six benchmarks.
The reason is that \ours {} belongs to \textit{class-conditional distribution} induced augmentation, while others tend to generate data through \textit{instance pairs} across domains.

To better demonstrate the effectiveness of \ours, we also reported the results of \ours {} with larger batch size following the setting in RDM \cite{nguyen2023domain}, namely \ours-L.
Compared to \ours {} and other SOTAs, \ours-L achieves significant performances with the improvement of 0.9\% on average, and has impressive performance on each benchmark.

\begin{figure*}[t]
	\centering
	\includegraphics[width=0.98\linewidth]{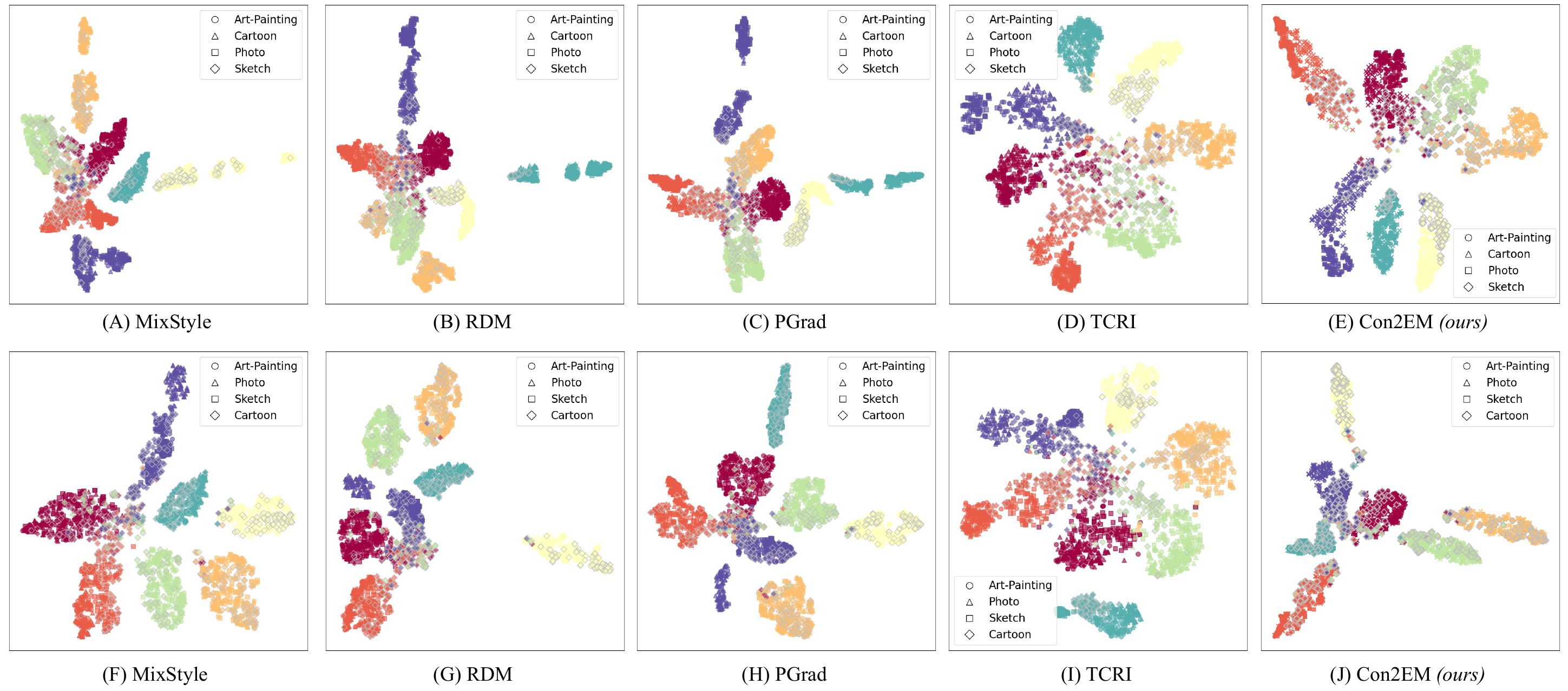}
	\caption{Visualization of learned representations on PACS using t-SNE. Different colors correspond to different classes and different shapes correspond to different domains. (A)-(E) and (F)-(J) select Sketch and Cartoon as target domain, respectively.}
	\label{fig:exp:tsne}
\end{figure*}

\subsection{Ablation and Parameter Studies}

Tab. \ref{tab:ablation} reported our ablation study using PACS as an example.
We can find that without regularization of distribution-level classifier, i.e., w/o $\mathcal{L}_{dis}$, the calculated distribution information becomes unstable, leading to performance degradation.
w/o $\hat{P}^{aug}$ means the exclusion of the smoothness term, effectively discarding Eq. \eqref{eq:mixup}.
Therefore, compared baseline and w/o $\hat{P}^{aug}$, we can find that the instances sampling from the generated domain-related class-conditional distributions can enhance generalizability.
Moreover, compared to the last two rows, we can find that smoothness can further improve performance in the case of limited domain.

\begin{table}[t]
	\centering
	\caption{Ablation study on PACS. w/o is short for without.}
	\label{tab:ablation}
	\tabcolsep=0.4em
	\begin{tabular}{l lllll}
		\toprule
		{} & A & C & P & S & Avg \\
		\midrule
		baseline                  & 84.7                  & 80.8                  & 97.2                  & 79.3                  & 85.5 \\
		w/o $\mathcal{L}_{dis}$   & 85.2 {\tiny ($+0.5$)} & 80.4 {\tiny ($-0.4$)} & 97.1 {\tiny ($-0.1$)} & 75.8 {\tiny ($-3.5$)} & 84.6 {\tiny ($-0.9$)} \\
		w/o $\hat{P}^{aug}$       & 86.4 {\tiny ($+1.7$)} & 81.6 {\tiny ($+0.8$)} & 97.5 {\tiny ($+0.3$)} & 81.0 {\tiny ($+1.7$)} & 86.6 {\tiny ($+1.1$)} \\
		\ours                     & 87.6 {\tiny ($+2.9$)} & 82.2 {\tiny ($+1.4$)} & 97.7 {\tiny ($+0.5$)} & 81.5 {\tiny ($+2.2$)} & 87.3 {\tiny ($+1.8$)} \\
		\bottomrule
	\end{tabular}
\end{table}

\begin{table}[t]
	\centering
	\caption{The computational cost (s) per batch compared to representative SOTAs on representative Benchmarks.}
	\label{tab:timecost}
	\begin{tabular}{l ccccc}
		\toprule
		{} & SAGM & RDM & PGrad & TCRI & \ours~(\textit{ours}) \\
		\midrule
		PACS & 0.403 & 0.201 & 0.953 & 0.432 & 0.222 \\
		OfficeHome & 0.412 & 0.210 & 0.982 & 0.445 & 0.276 \\
		DomainNet & 0.605 & 0.320 & 1.411 & 0.623 & 0.487 \\
		\bottomrule
	\end{tabular}
\end{table}

From Fig. \ref{fig:exp:tsne}, we can discover that among all approaches, the latent representations within each individual classes can distinguish the domain label to some extent.
Therefore, according to Subfig. (B), (D), (G), and (I), we can empirically demonstrate that aligning marginal distributions does not necessarily align conditional distributions as well.
Then, compared to other representative approaches, the representations of target domain in our \ours {} have not gathered together.
This phenomenon is particularly evident when Sketch is selected as the target domain.
What's more, compared to the others, each class exhibits good distinguishability between its latent represnetations in our \ours.
These observations directly validate our viewpoints, where acquiring discriminative generalization between classes within domains is more important than the alignment distribution.

\begin{figure}[t]
	\centering
	\includegraphics[width=0.9\linewidth]{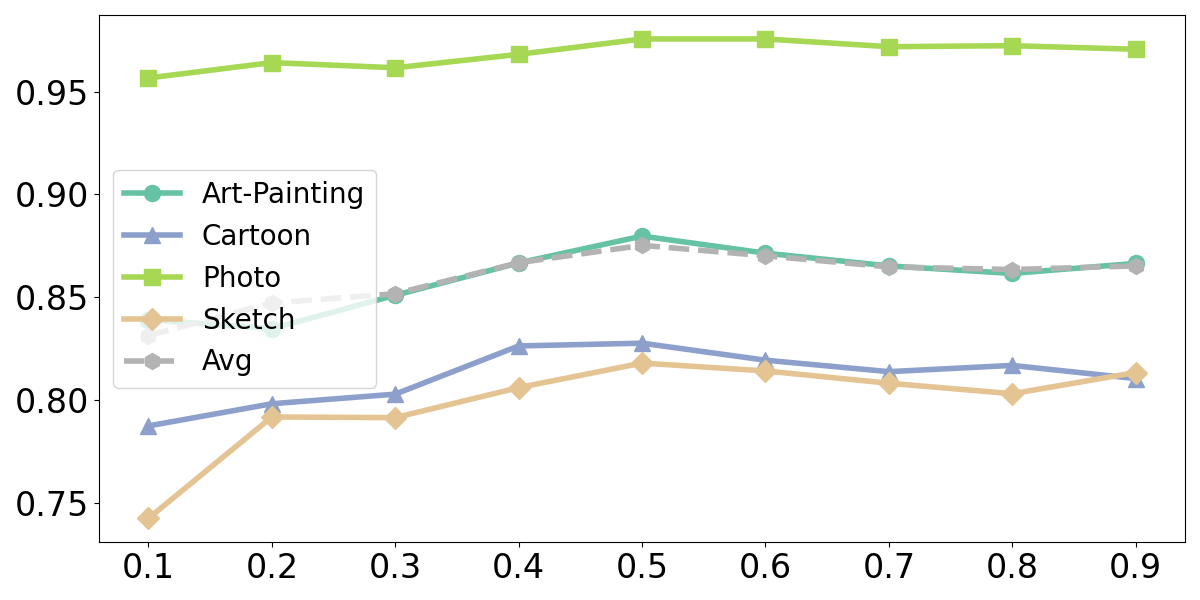}
	\caption{The influence of the trade-off $\lambda$.}
	\label{fig:exp:lambda}
\end{figure}

Fig. \ref{fig:exp:lambda} reported the results of varying the trade-off $\lambda$, which modulates the discriminability of generated distributions.
As $\lambda$ increases, the generated distributions closely resemble the observed source domains, leading to the trivial solution.
Conversely, a decrease in $\lambda$ results in more ambiguous semantics, leading to poor generalizability.
Therefore, $\lambda$ has been fixed to 0.5 during each training phase.

Tab. \ref{tab:timecost} reported the time costs of representative approaches on representative Benchmarks.
We can find that our method is faster than almost all SOTAs while maintaining the effectiveness.
This advantage arises because SAGM and PGrad require gradient manipulation, while TCRI must decouple representations for each class by HSIC with a kernel trick and utilize an additional larger classifier for each domain.
Therefore, compared to SAGM, PGrad and TCRI, our \ours {} can demonstrate its superiority in both performance and time cost.
Additionally, while RDM incurs lower time costs, our approach shows significantly better performance.

\section{Conclusion} \label{sec:conclusion}

In this paper, according to the phenomenon, where domain-related peaks or clusters probably emerge within each individual classes, we shift from the domain angle to the class angle for Domain Generalization.
In this way, rather than focusing on the alignment distribution, we can treat these domain-related clusters, i.e., class-conditional distributions, as hyper-instances.
The goal is to maintain their diversity within the distribution space, akin to conventional classification of individual instances.
Therefore, we can sample the instances from the semantically generated class-conditional distributions to provide feedback to the primordial instance-level classifier.
To achieve this goal, we design a novel Conjugate Consistent Enhanced Module, namely \ours.
Specifically, a distribution-level Universum strategy has been employed to generate such mentioned distributions through mixing the distributions from low-density areas with a particular semantic distribution.
To ensure generation accuracy, an additional distribution-level classifier has been adopted with the kernel trick.

\appendix

All results on six benchmarks were reported as following:

\begin{table}[ht]
	\centering
	\caption{Overall results on C-MNIST Benchmark. The bold and underline items are the best and the second-best results, respectively.}
	\label{tab:app_cmnist}
	\begin{tabular}{lcccc}
		\toprule
		\textbf{Algorithm} & \textbf{+90\%} & \textbf{+80\%} & \textbf{-90\%} & \textbf{Avg} \\
		\midrule
		ERM                  & 71.7 $\pm$ 0.1       & 72.9 $\pm$ 0.2       & 10.0 $\pm$ 0.1       & 51.5                 \\
		Mixup                & 72.7 $\pm$ 0.4       & 73.4 $\pm$ 0.1       & 10.1 $\pm$ 0.1       & 52.1                 \\
		DANN                 & 71.4 $\pm$ 0.9       & 73.1 $\pm$ 0.1       & 10.0 $\pm$ 0.0       & 51.5                 \\
		MLDG                 & 71.5 $\pm$ 0.2       & 73.1 $\pm$ 0.2       & 9.8 $\pm$ 0.1        & 51.5                 \\
		GroupDRO             & \textbf{73.1 $\pm$ 0.3}       & 73.2 $\pm$ 0.2       & 10.0 $\pm$ 0.2       & 52.1                 \\
		IRM                  & 72.5 $\pm$ 0.1       & 73.3 $\pm$ 0.5       & 10.2 $\pm$ 0.3       & 52.0                 \\
		ARM                  & 72.7 $\pm$ 0.6       & 73.0 $\pm$ 0.8       & 10.0 $\pm$ 0.1       & 51.9\\
		VREx                 & 72.4 $\pm$ 0.3       & 72.9 $\pm$ 0.4       & 10.2 $\pm$ 0.0       & 51.8                 \\
		EQRM                 & 72.4 $\pm$ 0.5       & \underline{73.5 $\pm$ 0.1}       & 10.1 $\pm$ 0.1       & 52.0                 \\
		Fish                 & 71.0 $\pm$ 0.2       & 73.0 $\pm$ 0.1       & 10.2 $\pm$ 0.1       & 51.4                 \\
		Fishr                & 72.0 $\pm$ 0.4       & 73.4 $\pm$ 0.2       & 10.0 $\pm$ 0.1       & 51.8                 \\
		CORAL                & 71.6 $\pm$ 0.3       & 73.1 $\pm$ 0.1       & 9.9 $\pm$ 0.1        & 51.5                 \\
		MMD                  & 71.4 $\pm$ 0.3       & 73.1 $\pm$ 0.2       & 9.9 $\pm$ 0.3        & 51.5                 \\
		DANN                 & 71.4 $\pm$ 0.9       & 73.1 $\pm$ 0.1       & 10.0 $\pm$ 0.0       & 51.5                 \\
		SagNet               & 71.8 $\pm$ 0.2       & 73.0 $\pm$ 0.2       & \underline{10.3 $\pm$ 0.0}       & 51.7                 \\
		Mixstyle             & 72.8 $\pm$ 0.4       & 73.2 $\pm$ 0.2       & 9.9 $\pm$ 0.0        & 52.0 \\
		RDM                  & \underline{73.0 $\pm$ 0.3}       & 73.3 $\pm$ 0.2       & 10.1 $\pm$ 0.1       & \underline{52.2}                 \\
		SAGM                 & 71.7 $\pm$ 0.6       & 73.2 $\pm$ 0.2       & 10.2 $\pm$ 0.2       & 51.7                 \\
		PGard                & 72.2 $\pm$ 0.1       & 73.4 $\pm$ 0.3       & 10.1 $\pm$ 0.1       & 51.9                 \\
		TCRI                 & 72.8 $\pm$ 0.3       & \underline{73.5 $\pm$ 0.6}       & \underline{10.3 $\pm$ 0.1}       & 52.1                 \\
		\midrule
		\ours                & 72.3 $\pm$ 0.1       & \textbf{73.6 $\pm$ 0.3}       & \textbf{10.8 $\pm$ 0.4}       & \textbf{52.3}                 \\
		\ours-L                & 72.3 $\pm$ 0.1       & \textbf{73.6 $\pm$ 0.3}       & \textbf{10.8 $\pm$ 0.4}       & \textbf{52.3}                 \\
		\bottomrule
	\end{tabular}
\end{table}

\begin{table}[ht]
	\centering
	\tabcolsep=0.52em
	\caption{Overall results on VLCS Benchmark.}
	\label{tab:app_vlcs}
	\begin{tabular}{lccccc}
		\toprule 
		\textbf{Algorithm} & \textbf{C} & \textbf{L} & \textbf{S} & \textbf{V} & \textbf{Avg} \\
		\midrule 
		ERM & 97.7 $\pm$ 0.4 & 64.3 $\pm$ 0.9 & 73.4 $\pm$ 0.5 & 74.6 $\pm$ 1.3 & 77.5 \\
		Mixup & 98.3 $\pm$ 0.6 & 64.8 $\pm$ 1.0 & 72.1 $\pm$ 0.5 & 74.3 $\pm$ 0.8 & 77.4 \\
		DANN & 99.0 $\pm$ 0.3 & 65.1 $\pm$ 1.4 & 73.1 $\pm$ 0.3 & 77.2 $\pm$ 0.6 & 78.6 \\
		MLDG & 97.4 $\pm$ 0.2 & 65.2 $\pm$ 0.7 & 71.0 $\pm$ 1.4 & 75.3 $\pm$ 1.0 & 77.2 \\
		GroupDRO & 97.3 $\pm$ 0.3 & 63.4 $\pm$ 0.9 & 69.5 $\pm$ 0.8 & 76.7 $\pm$ 0.7 & 76.7 \\
		IRM & 98.6 $\pm$ 0.1 & 64.9 $\pm$ 0.9 & 73.4 $\pm$ 0.6 & 77.3 $\pm$ 0.9 & 78.5 \\
		ARM & 98.7 $\pm$ 0.2 & 63.6 $\pm$ 0.7 & 71.3 $\pm$ 1.2 & 76.7 $\pm$ 0.6 & 77.6 \\
		VREx & 98.4 $\pm$ 0.3 & 64.4 $\pm$ 1.4 & 74.1 $\pm$ 0.4 & 76.2 $\pm$ 1.3 & 78.3 \\
		EQRM & 98.3 $\pm$ 0.0 & 63.7 $\pm$ 0.8 & 72.6 $\pm$ 1.0 & 76.7 $\pm$ 1.1 & 77.8 \\
		Fish & - & - & - & - & 77.8 \\
		Fishr & 98.9 $\pm$ 0.3 & 64.0 $\pm$ 0.5 & 71.5 $\pm$ 0.2 & 76.8 $\pm$ 0.7 & 77.8 \\
		CORAL & 98.3 $\pm$ 0.1 & 66.1 $\pm$ 1.2 & 73.4 $\pm$ 0.3 & 77.5 $\pm$ 1.2 & 78.8 \\
		MMD & 97.7 $\pm$ 0.1 & 64.0 $\pm$ 1.1 & 72.8 $\pm$ 0.2 & 75.3 $\pm$ 3.3 & 77.5 \\
		SagNet & 97.9 $\pm$ 0.4 & 64.5 $\pm$ 0.5 & 71.4 $\pm$ 1.3 & 77.5 $\pm$ 0.5 & 77.8 \\
		Mixstyle & 98.6 $\pm$ 0.3 & 64.5 $\pm$ 1.1 & 72.6 $\pm$ 0.5 & 75.7 $\pm$ 1.7 & 77.9 \\
		RDM & 98.1 $\pm$ 0.2 & 64.9 $\pm$ 0.7 & 72.6 $\pm$ 0.5 & 77.9 $\pm$ 1.2 & 78.4 \\
		SAGM & \underline{99.0 $\pm$ 0.2} & 65.2 $\pm$ 0.4 & \textbf{75.1} $\pm$ 0.3 & \textbf{80.7 $\pm$ 0.8} & {80.0} \\
		PGrad & \textbf{99.1 $\pm$ 0.1} & 63.8 $\pm$ 0.7 & 73.5 $\pm$ 0.5 & 79.0 $\pm$ 0.5 & 78.9 \\
		TCRI & \textbf{99.1 $\pm$ 0.4} & \underline{67.2 $\pm$ 0.6} & 74.4 $\pm$ 0.6 & 79.0 $\pm$ 0.6 & 79.9 \\
		\midrule 
		\ours & 98.9 $\pm$ 0.4 & \textbf{67.5 $\pm$ 0.5} & 74.7 $\pm$ 0.6 & 79.4 $\pm$ 0.6 & \underline{80.2} \\
		\ours-L & \underline{99.0 $\pm$ 0.3} & \textbf{67.5 $\pm$ 0.2} & \underline{75.0 $\pm$ 0.4} & \underline{79.5 $\pm$ 0.5} & \textbf{80.3} \\
		\bottomrule
	\end{tabular}
\end{table}

\begin{table}[ht]
	\centering
	\tabcolsep=0.52em
	\caption{Overall results on PACS Benchmark.}
	\label{tab:app_pacs}
	\begin{tabular}{lccccc}
		\toprule 
		\textbf{Algorithm} & \textbf{A} & \textbf{C} & \textbf{P} & \textbf{S} & \textbf{Avg} \\
		\midrule
		ERM & 84.7 $\pm$ 0.4 & 80.8 $\pm$ 0.6 & 97.2 $\pm$ 0.3 & 79.3 $\pm$ 1.0 & 85.5 \\
		Mixup & 86.1 $\pm$ 0.5 & 78.9 $\pm$ 0.8 & 97.6 $\pm$ 0.1 & 75.8 $\pm$ 1.8 & 84.6 \\
		DANN & 86.4 $\pm$ 0.8 & 77.4 $\pm$ 0.8 & 97.3 $\pm$ 0.4 & 73.5 $\pm$ 2.3 & 83.6 \\
		MLDG & 85.5 $\pm$ 1.4 & 80.1 $\pm$ 1.7 & 97.4 $\pm$ 0.3 & 76.6 $\pm$ 1.1 & 84.9 \\
		GroupDRO & 83.5 $\pm$ 0.9 & 79.1 $\pm$ 0.6 & 96.7 $\pm$ 0.3 & 78.3 $\pm$ 2.0 & 84.4 \\
		IRM & 84.8 $\pm$ 1.3 & 76.4 $\pm$ 1.1 & 96.7 $\pm$ 0.6 & 76.1 $\pm$ 1.0 & 83.5 \\
		ARM & 86.8 $\pm$ 0.6 & 76.8 $\pm$ 0.5 & 97.4 $\pm$ 0.3 & 79.3 $\pm$ 1.2 & 85.1 \\
		VREx & 86.0 $\pm$ 1.6 & 79.1 $\pm$ 0.6 & 96.9 $\pm$ 0.5 & 77.7 $\pm$ 1.7 & 84.9 \\
		EQRM & 86.5 $\pm$ 0.4 & {82.1 $\pm$ 0.7} & 96.6 $\pm$ 0.2 & 80.8 $\pm$ 0.2 & 86.5 \\
		Fish & - & - & - & - & 85.5 \\
		Fishr & 88.4 $\pm$ 0.2 & 78.7 $\pm$ 0.7 & 97.0 $\pm$ 0.1 & 77.8 $\pm$ 2.0 & 85.5 \\
		CORAL & 88.3 $\pm$ 0.2 & 80.0 $\pm$ 0.5 & 97.5 $\pm$ 0.3 & 78.8 $\pm$ 1.3 & 86.2 \\
		MMD & 86.1 $\pm$ 1.4 & 79.4 $\pm$ 0.9 & 96.6 $\pm$ 0.2 & 76.5 $\pm$ 0.5 & 84.6 \\
		SagNet & 87.4 $\pm$ 1.0 & 80.7 $\pm$ 0.6 & 97.1 $\pm$ 0.1 & 80.0 $\pm$ 0.4 & 86.3 \\
		Mixstyle & 86.8 $\pm$ 0.5 & 79.0 $\pm$ 1.4 & 96.9 $\pm$ 0.1 & 78.5 $\pm$ 2.3 & 85.2 \\
		RDM & {88.4 $\pm$ 0.2} & 81.3 $\pm$ 1.6 & 97.1 $\pm$ 0.1 & \textbf{81.8 $\pm$ 1.1} & {87.2} \\
		SAGM & 87.4 $\pm$ 0.2 & 80.2 $\pm$ 0.3 & \underline{98.0 $\pm$ 0.2} & 80.8 $\pm$ 0.6 & 86.6 \\
		PGrad & \textbf{89.6 $\pm$ 0.3} & 78.9 $\pm$ 0.6 & 97.7 $\pm$ 0.3 & 78.8 $\pm$ 1.0 & 86.2 \\
		TCRI & 87.8 $\pm$ 0.7 & 81.8 $\pm$ 0.6 & \textbf{98.1 $\pm$ 0.4} & 80.3 $\pm$ 0.4 &  87.0 \\
		\midrule
		\ours & 87.6 $\pm$ 1.0 & \underline{82.2 $\pm$ 0.3} & 97.7 $\pm$ 0.4 & {81.5 $\pm$ 0.6} & \underline{87.3} \\
		\ours-L & \underline{88.5 $\pm$ 0.7} & \textbf{82.3 $\pm$ 0.4} & 97.9 $\pm$ 0.2 & \underline{81.6 $\pm$ 0.6} & \textbf{87.6} \\
		\bottomrule
	\end{tabular}
\end{table}

\begin{table}[ht]
	\centering
	\tabcolsep=0.52em
	\caption{Overall results on OfficeHome Benchmark.}
	\label{tab:app_officehome}
	\begin{tabular}{lccccc}
		\toprule 
		\textbf{Algorithm}   & \textbf{A}           & \textbf{C}           & \textbf{P}           & \textbf{R}           & \textbf{Avg}         \\
		\midrule
		ERM                  & 61.3 $\pm$ 0.7       & 52.4 $\pm$ 0.3       & 75.8 $\pm$ 0.1       & 76.6 $\pm$ 0.3       & 66.5                 \\
		Mixup                & 62.4 $\pm$ 0.8       & 54.8 $\pm$ 0.6       & 76.9 $\pm$ 0.3       & 78.3 $\pm$ 0.2       & 68.1                 \\
		DANN                 & 59.9 $\pm$ 1.3       & 53.0 $\pm$ 0.3       & 73.6 $\pm$ 0.7       & 76.9 $\pm$ 0.5       & 65.9                 \\
		MLDG                 & 61.5 $\pm$ 0.9       & 53.2 $\pm$ 0.6       & 75.0 $\pm$ 1.2       & 77.5 $\pm$ 0.4       & 66.8                 \\
		GroupDRO             & 60.4 $\pm$ 0.7       & 52.7 $\pm$ 1.0       & 75.0 $\pm$ 0.7       & 76.0 $\pm$ 0.7       & 66.0                 \\		
		IRM                  & 58.9 $\pm$ 2.3       & 52.2 $\pm$ 1.6       & 72.1 $\pm$ 2.9       & 74.0 $\pm$ 2.5       & 64.3                 \\
		ARM                  & 58.9 $\pm$ 0.8       & 51.0 $\pm$ 0.5       & 74.1 $\pm$ 0.1       & 75.2 $\pm$ 0.3       & 64.8                 \\
		VREx                 & 60.7 $\pm$ 0.9       & 53.0 $\pm$ 0.9       & 75.3 $\pm$ 0.1       & 76.6 $\pm$ 0.5       & 66.4                 \\
		EQRM                 & 60.5 $\pm$ 0.1       & {56.0 $\pm$ 0.2}       & 76.1 $\pm$ 0.4       & 77.4 $\pm$ 0.3       & 67.5                 \\
		Fish                 & -                    & -                    & -                    & -                    & 68.6                 \\
		Fishr                & 62.4 $\pm$ 0.5       & 54.4 $\pm$ 0.4       & 76.2 $\pm$ 0.5       & 78.3 $\pm$ 0.1       & 67.8		            \\
		CORAL                & 65.3 $\pm$ 0.4       & 54.4 $\pm$ 0.5       & {76.5 $\pm$ 0.1}       & {78.4 $\pm$ 0.5}       & 68.7                 \\
		MMD                  & 60.4 $\pm$ 0.2       & 53.3 $\pm$ 0.3       & 74.3 $\pm$ 0.1       & 77.4 $\pm$ 0.6       & 66.3                 \\
		SagNet               & 63.4 $\pm$ 0.2       & 54.8 $\pm$ 0.4       & 75.8 $\pm$ 0.4       & 78.3 $\pm$ 0.3       & 68.1                 \\
		Mixstyle             & 51.1 $\pm$ 0.3       & 53.2 $\pm$ 0.4       & 68.2 $\pm$ 0.7       & 69.2 $\pm$ 0.6       & 60.4                 \\
		RDM                  & 61.1 $\pm$ 0.4       & 55.1 $\pm$ 0.3       & 75.7 $\pm$ 0.5       & 77.3 $\pm$ 0.3       & 67.3                 \\
		SAGM                 & {65.4 $\pm$ 0.4}       & \textbf{57.0 $\pm$ 0.3}       & \underline{78.0 $\pm$ 0.3}       & \textbf{80.0 $\pm$ 0.2}       & \textbf{70.1}        \\
		PGrad                & \textbf{65.8 $\pm$ 0.2}       & 55.4 $\pm$ 0.4       & \underline{78.0 $\pm$ 0.1}       & \textbf{80.0 $\pm$ 0.4}       & {69.8}     \\
		TCRI                 & 62.9 $\pm$ 0.6       & 54.0 $\pm$ 0.0       & {76.5 $\pm$ 0.5}       & 78.0 $\pm$ 0.4       & 67.9                 \\
		\midrule 
		\ours                & 61.8 $\pm$ 0.8       & 54.0 $\pm$ 0.5       & 76.3 $\pm$ 0.3       & 78.0 $\pm$ 0.3       & 67.5                 \\
		\ours-L              & \underline{65.7 $\pm$ 0.6}       & \underline{56.3 $\pm$ 0.5}       & \textbf{78.2 $\pm$ 0.2}       & \underline{79.8 $\pm$ 0.1}       & \underline{70.0}                 \\
		\bottomrule
	\end{tabular}
\end{table}

\begin{table}[ht]
	\centering
	\tabcolsep=0.52em
	\caption{Overall results on TerraIncognita Benchmark.}
	\label{tab:app_terraincognita}
	\begin{tabular}{lccccc}
		\toprule 
		\textbf{Algorithm} & \textbf{L100} & \textbf{L38} & \textbf{L43} & \textbf{L46} & \textbf{Avg} \\
		\midrule 
		ERM & 49.8 $\pm$ 4.4 & 42.1 $\pm$ 1.4 & 56.9 $\pm$ 1.8 & 35.7 $\pm$ 3.9 & 46.1 \\
		Mixup & \underline{59.6 $\pm$ 2.0} & 42.2 $\pm$ 1.4 & 55.9 $\pm$ 0.8 & 33.9 $\pm$ 1.4 & 47.9 \\
		DANN & 51.1 $\pm$ 3.5 & 40.6 $\pm$ 0.6 & 57.4 $\pm$ 0.5 & 37.7 $\pm$ 1.8 & 46.7 \\
		MLDG & 54.2 $\pm$ 3.0 & 44.3 $\pm$ 1.1 & 55.6 $\pm$ 0.3 & 36.9 $\pm$ 2.2 & 47.7 \\
		GroupDRO & 41.2 $\pm$ 0.7 & 38.6 $\pm$ 2.1 & 56.7 $\pm$ 0.9 & 36.4 $\pm$ 2.1 & 43.2 \\
		IRM & 54.6 $\pm$ 1.3 & 39.8 $\pm$ 1.9 & 56.2 $\pm$ 1.8 & 39.6 $\pm$ 0.8 & 47.6 \\
		ARM & 49.3 $\pm$ 0.7 & 38.3 $\pm$ 2.4 & 55.8 $\pm$ 0.8 & 38.7 $\pm$ 1.3 & 45.5 \\
		VREx & 48.2 $\pm$ 4.3 & 41.7 $\pm$ 1.3 & 56.8 $\pm$ 0.8 & 38.7 $\pm$ 3.1 & 46.4 \\
		EQRM & 47.9 $\pm$ 1.9 & 45.2 $\pm$ 0.3 & 59.1 $\pm$ 0.3 & 38.8 $\pm$ 0.6 & 47.8 \\
		Fish & - & - & - & - & 45.1 \\
		Fishr & 50.2 $\pm$ 3.9 & 43.9 $\pm$ 0.8 & 55.7 $\pm$ 2.2 & 39.8 $\pm$ 1.0 & 47.4 \\
		CORAL & 51.6 $\pm$ 2.4 & 42.2 $\pm$ 1.0 & 57.0 $\pm$ 1.0 & 39.8 $\pm$ 2.9 & 47.6 \\
		MMD & 41.9 $\pm$ 3.0 & 34.8 $\pm$ 1.0 & 57.0 $\pm$ 1.9 & 35.2 $\pm$ 1.8 & 42.2 \\
		SagNet & 53.0 $\pm$ 2.9 & 43.0 $\pm$ 2.5 & 57.9 $\pm$ 0.6 & 40.4 $\pm$ 1.3 & 48.6 \\
		Mixstyle & 54.3 $\pm$ 1.1 & 34.1 $\pm$ 1.1 & 55.9 $\pm$ 1.1 & 31.7 $\pm$ 2.1 & 44.0 \\
		RDM & 52.9 $\pm$ 1.2 & 43.1 $\pm$ 1.0 & 58.1 $\pm$ 1.3 & 36.1 $\pm$ 2.9 & 47.5 \\
		SAGM & 54.8 $\pm$ 1.3 & 41.4 $\pm$ 0.8 & 57.7 $\pm$ 0.6 & 41.3 $\pm$ 0.4 & 48.8 \\
		PGrad & \textbf{61.2 $\pm$ 2.2} & {45.7 $\pm$ 0.9} & 58.2 $\pm$ 0.3 & 37.9 $\pm$ 0.7 & 50.7 \\
		TCRI & 58.9 $\pm$ 1.6 & 45.0 $\pm$ 1.0 & \textbf{59.3 $\pm$ 0.2} & \underline{41.5 $\pm$ 0.5} & \underline{51.2} \\
		\midrule 
		\ours & 59.2 $\pm$ 3.0 & \underline{46.2 $\pm$ 0.8} & \underline{59.2 $\pm$ 0.2} & \textbf{43.6 $\pm$ 1.4} & \underline{52.1} \\
		\ours-L & 59.4 $\pm$ 2.8 & \textbf{46.3 $\pm$ 0.7} & \textbf{59.3 $\pm$ 0.3} & \textbf{43.6 $\pm$ 1.0} & \textbf{52.2} \\
		\bottomrule
	\end{tabular}
\end{table}

\begin{table*}[ht]
	\centering
	\caption{Overall results on DomainNet Benchmark.}
	\label{tab:app_domainnet}
	\begin{tabular}{lccccccc}
		\toprule
		\textbf{Algorithm} & \textbf{C} & \textbf{I} & \textbf{P} & \textbf{Q} & \textbf{R} & \textbf{S} & \textbf{Avg} \\
		\midrule
		ERM & 58.1 $\pm$ 0.3 & 18.8 $\pm$ 0.3 & 46.7 $\pm$ 0.3 & 12.2 $\pm$ 0.4 & 59.6 $\pm$ 0.1 & 49.8 $\pm$ 0.4 & 40.9 \\
		Mixup & 55.7 $\pm$ 0.3 & 18.5 $\pm$ 0.5 & 44.3 $\pm$ 0.5 & 12.5 $\pm$ 0.4 & 55.8 $\pm$ 0.3 & 48.2 $\pm$ 0.5 & 39.2 \\
		DANN & 53.1 $\pm$ 0.2 & 18.3 $\pm$ 0.1 & 44.2 $\pm$ 0.7 & 11.8 $\pm$ 0.1 & 55.5 $\pm$ 0.4 & 46.8 $\pm$ 0.6 & 38.3 \\
		MLDG & 59.1 $\pm$ 0.2 & 19.1 $\pm$ 0.3 & 45.8 $\pm$ 0.7 & 13.4 $\pm$ 0.3 & 59.6 $\pm$ 0.1 & 50.2 $\pm$ 0.4 & 41.2 \\
		GroupDRO & 47.2 $\pm$ 0.5 & 17.5 $\pm$ 0.4 & 33.8 $\pm$ 0.5 & 9.3 $\pm$ 0.3 & 51.6 $\pm$ 0.4 & 40.1 $\pm$ 0.6 & 33.3 \\
		IRM & 48.5 $\pm$ 2.8 & 15.0 $\pm$ 1.5 & 38.3 $\pm$ 4.3 & 10.9 $\pm$ 0.5 & 48.2 $\pm$ 5.2 & 42.3 $\pm$ 3.1 & 33.9 \\
		ARM & 49.7 $\pm$ 0.3 & 16.3 $\pm$ 0.5 & 40.9 $\pm$ 1.1 & 9.4 $\pm$ 0.1 & 53.4 $\pm$ 0.4 & 43.5 $\pm$ 0.4 & 35.5 \\
		VREx & 47.3 $\pm$ 3.5 & 16.0 $\pm$ 1.5 & 35.8 $\pm$ 4.6 & 10.9 $\pm$ 0.3 & 49.6 $\pm$ 4.9 & 42.0 $\pm$ 3.0 & 33.6 \\
		EQRM & 56.1 $\pm$ 1.3 & 19.6 $\pm$ 0.1 & 46.3 $\pm$ 1.5 & 12.9 $\pm$ 0.3 & 61.1 $\pm$ 0.0 & 50.3 $\pm$ 0.1 & 41.0 \\
		Fish & - & - & - & - & - & - & 42.7 \\
		Fishr & 58.2 $\pm$ 0.5 & 20.2 $\pm$ 0.2 & 47.7 $\pm$ 0.3 & 12.7 $\pm$ 0.2 & 60.3 $\pm$ 0.2 & 50.8 $\pm$ 0.1 & 41.7 \\
		CORAL & 59.2 $\pm$ 0.1 & 19.7 $\pm$ 0.2 & 46.6 $\pm$ 0.3 & 13.4 $\pm$ 0.4 & 59.8 $\pm$ 0.2 & 50.1 $\pm$ 0.6 & 41.5 \\
		MMD & 32.1 $\pm$ 13.3 & 11.0 $\pm$ 4.6 & 26.8 $\pm$ 11.3 & 8.7 $\pm$ 2.1 & 32.7 $\pm$ 13.8 & 28.9 $\pm$ 11.9 & 23.4 \\
		SagNet & 57.7 $\pm$ 0.3 & 19.0 $\pm$ 0.2 & 45.3 $\pm$ 0.3 & 12.7 $\pm$ 0.5 & 58.1 $\pm$ 0.5 & 48.8 $\pm$ 0.2 & 40.3 \\
		Mixstyle & 51.9 $\pm$ 0.4 & 13.3 $\pm$ 0.2 & 37.0 $\pm$ 0.5 & 12.3 $\pm$ 0.1 & 46.1 $\pm$ 0.3 & 43.4 $\pm$ 0.4 & 34.0 \\
		RDM & 62.1 $\pm$ 0.2 & 20.7 $\pm$ 0.1 & 49.2 $\pm$ 0.4 & 14.1 $\pm$ 0.2 & 63.0 $\pm$ 1.3 & 51.4 $\pm$ 0.1 & 43.4 \\
		SAGM & \underline{64.9 $\pm$ 0.2} & {21.1 $\pm$ 0.3} & \underline{51.5 $\pm$ 0.2} & {14.8 $\pm$ 0.2} & \textbf{64.1 $\pm$ 0.2} & \textbf{53.6 $\pm$ 0.2} & \underline{45.0} \\
		PGrad & 57.0 $\pm$ 0.5 & 18.2 $\pm$ 0.2 & 48.4 $\pm$ 0.2 & 13.0 $\pm$ 0.1 & 60.9 $\pm$ 0.1 & 48.8 $\pm$ 0.1 & 41.0 \\
		TCRI & 58.6 $\pm$ 0.3 & 18.6 $\pm$ 0.1 & 49.4 $\pm$ 0.2 & 12.9 $\pm$ 0.1 & 61.5 $\pm$ 0.2 & 50.8 $\pm$ 0.1 & 42.2 \\
		\midrule
		\ours & 64.6 $\pm$ 0.4 & \underline{22.0 $\pm$ 0.3} & 50.8 $\pm$ 0.4 & \underline{14.9 $\pm$ 0.2} & 63.6 $\pm$ 0.3 & 52.6 $\pm$ 0.2 & {44.8} \\
		\ours-L & \textbf{65.0 $\pm$ 0.3} & \textbf{22.4 $\pm$ 0.2} & \textbf{51.8 $\pm$ 0.1} & \textbf{15.0 $\pm$ 0.2} & \underline{63.6 $\pm$ 0.2} & \underline{53.5 $\pm$ 0.2} & \textbf{45.2} \\
		\bottomrule
	\end{tabular}
\end{table*}

\bibliographystyle{IEEEtran}
\bibliography{ref}

\newpage

\section{Biography Section}
 
\vspace{11pt}

\begin{IEEEbiography}[{\includegraphics[width=1in,height=1.25in,clip,keepaspectratio]{figs/meng\_cao}}]{Meng Cao}
	received his B.S. degree in computer science from the Nanjing University of Information Science and Technology (NUIST), Nanjing, China, in 2017. In 2020, he has completed his M.S. degree in software engineering at NUIST. He is currently a Ph.D candidate at Nanjing University of Aeronautics and Astronautics (NUAA), and is a a member of ParNec Group, led by Professor Songcan Chen. His research interest is Domain Generalization.
\end{IEEEbiography}

\vspace{11pt}

\begin{IEEEbiography}[{\includegraphics[width=1in,height=1.25in,clip,keepaspectratio]{figs/songcan\_chen}}]{Songcan Chen}
	received the B.S. degree from Hangzhou University (now merged into Zhejiang University), Hangzhou, China, in 1983, the M.S. degree from Shanghai Jiao Tong University, Shanghai, China, in 1985, and the Ph.D. degree from the Nanjing University of Aeronautics and Astronautics (NUAA), Nanjing, China, in 1997. He joined NUAA in 1986, where he has been a full-time Professor with the Department of Computer Science and Engineering since 1998. He has authored/coauthored about 200 peer-reviewed scientific articles. His current research interests include pattern recognition, machine learning, and neural computing. Dr. Chen is also a fellow of the International Association of Pattern Recognition (IAPR) and the China Association of Artificial Intelligence (CAAI). He obtained Honorable Mentions of the 2006, 2007, and 2010 Best Paper Awards of Pattern Recognition journal.
\end{IEEEbiography}

\vfill

\end{document}